\documentclass[sigconf]{acmart}

\usepackage{placeins}
\usepackage{graphicx}
\usepackage{subcaption}
\usepackage{booktabs}
\usepackage{bm} 
\usepackage[table]{xcolor}
\usepackage{colortbl}
\usepackage[ruled,vlined]{algorithm2e}
\usepackage{enumitem}
\usepackage{float}
\usepackage{xcolor}
\usepackage{multirow} 
\usepackage{balance}
\definecolor{ShadowBG}{RGB}{230,230,230}   
\definecolor{HeaderGray}{RGB}{210, 210, 210}
\definecolor{BlockGray}{RGB}{230, 230, 230}
\definecolor{thirdcolor}{RGB}{18, 78, 34}
\definecolor{bestcolor}{RGB}{130, 53, 25} 
\definecolor{secondcolor}{RGB}{0, 0, 0}

\newcommand{\bg}[1]{\colorbox{ShadowBG}{#1}}
\begin{document}
\title{Partially Observable Learning for Multi-Platform Dispatch Optimization}

\author{Fengming Yao}
\email{fy283@exeter.ac.uk}
\affiliation{
  \institution{University of Exeter}
  \city{Exeter}
  \country{UK}
}

\author{Man Luo}
\email{M.Luo@exeter.ac.uk}
\affiliation{
  \institution{University of Exeter}
  \city{Exeter}
  \country{UK}
}

\begin{abstract}

Instant delivery platforms have become a critical component of urban logistics, increasingly relying on crowdsourced couriers to fulfill highly dynamic orders. In real-world systems, couriers are not exclusive to a single platform and may concurrently serve multiple platforms, while each platform can only observe its own orders and couriers' interactions due to privacy and operational constraints. 
This results in a multi-platform dispatch environment with inherent partial observability. However, most existing works on dispatch optimization assume full courier observability and mandatory assignment acceptance, causing substantial performance degradation when deployed in realistic multi-platform settings. 
In this paper, we propose \textbf{POLO}, a partially observable multi-agent reinforcement learning framework for dispatching optimization in multi-platform instant delivery systems. 
POLO firstly models each platform-grid pair as an independent agent that learns dispatch policies solely from platform-local observations, aligning the learning process with real-world privacy and operational constraints. 
To support effective decision-making under incomplete and heterogeneous courier information, POLO introduces a novel attention-based policy representation that selectively aggregates inter-courier information. Moreover, we design a counterfactual reward shaping mechanism to mitigate the non-stationarity induced by joint actions across grids, leading to more stable and scalable learning. 
We develop a high-fidelity simulator to evaluate dispatch performance under varying numbers of platforms and system scales. Extensive experiments demonstrate that POLO consistently outperforms strong baselines in terms of platform revenue and courier travel efficiency, highlighting its robustness and effectiveness in realistic multi-platform settings. The implementation of both the simulator and the proposed model is available at https://github.com/yaofengming1999/polo-courier.git

\end{abstract}



\maketitle
\section{Introduction}

\begin{figure}
    \centering
    \includegraphics[width=0.90\linewidth]{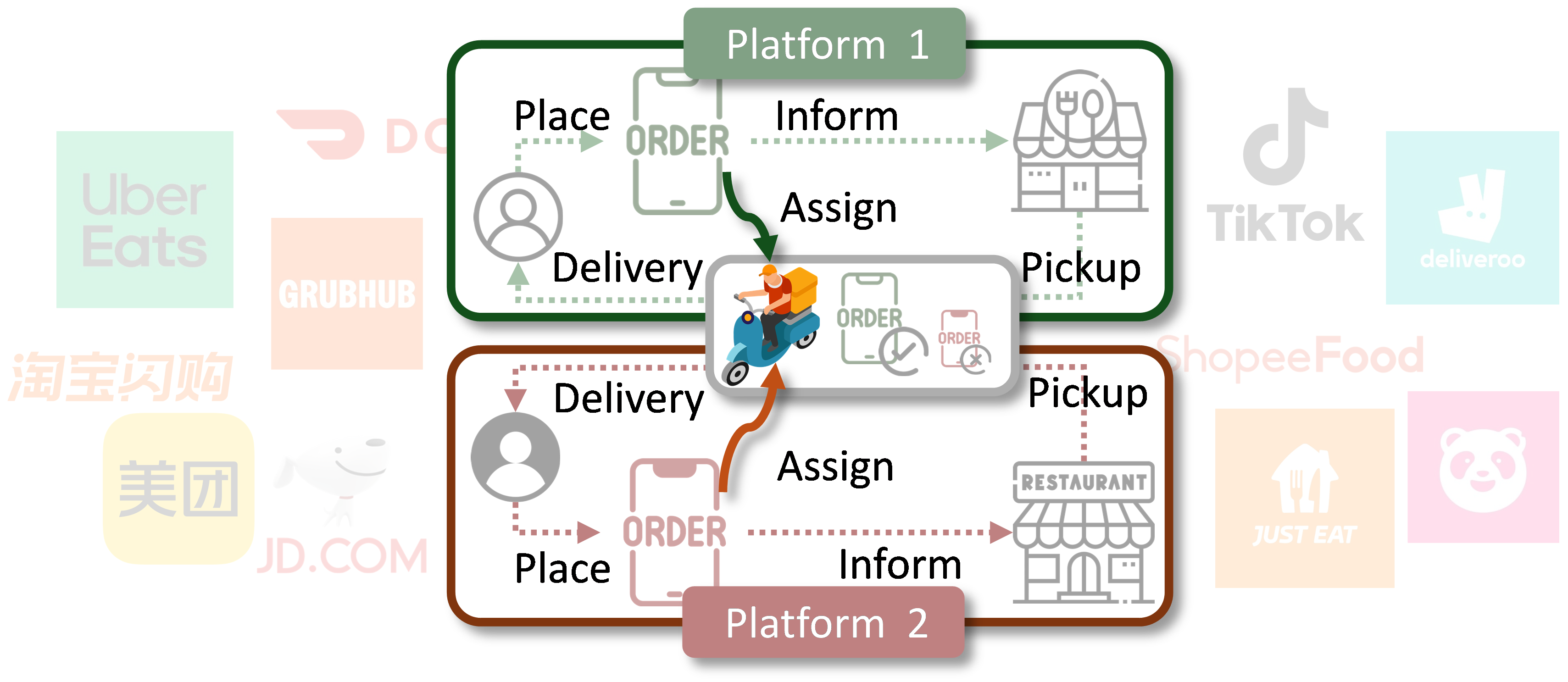}
    \caption{Multi-platform instant delivery. A courier may serve multiple platforms and reject the assigned order.}
    \label{fig:multiple_platform}
\end{figure}

Instant delivery provides on-demand transportation of goods, food, and groceries in response to real-time customer requests and has become a critical component of urban logistics systems~\cite{wen2024survey, dablanc2017rise}. 
The rapid growth of instant delivery markets has led to the coexistence of multiple delivery platforms within the same regions~\cite{dablanc2017rise, cennamo2021competing, tan2025research}, such as Meituan, Ele.me, and JD.com in China, Deliveroo, Uber Eats, and Just Eat in Europe, and DoorDash and Grubhub in the United States. This multi-platform setting fundamentally reshapes how dispatch decisions are made in urban delivery systems. 
What's more, couriers play a central role in this process, and a substantial fraction of them are crowdsourced workers who are not exclusively affiliated with a single platform~\cite{dablanc2017rise, alnaggar2021crowdsourced}. While crowdsourced couriers provide platforms with a flexible and scalable workforce~\cite{alnaggar2024compensation}, they also introduce limited control over courier behavior and increased uncertainty in system dynamics~\cite{savelsbergh2022challenges}. As illustrated in Fig.~\ref{fig:multiple_platform}, multi-platform dispatch is characterized by partial observability and decentralized control. Each platform can only observe its own orders and courier interactions, while courier availability, future routes, and acceptance decisions are jointly influenced by actions taken across competing platforms. As a result, the global system state is not accessible to any single decision-maker. This partial observability introduces strong non-stationarity into the order dispatch process and significantly degrades the performance of existing methods when applied to multi-platform settings. 

%
Order dispatch is a core decision-making problem in instant delivery and has been extensively studied~\cite{zong2025deep,shi2019operating}. Dispatch decisions directly affect routing feasibility, delivery timelines, customer satisfaction, and platform revenue~\cite{jiang2023faircod, wang2025coopride, yang2024rethinking, li2019efficient, tang2021value}. In practice, due to orders overlapping along courier routes, dispatch is naturally formulated as a concurrent spatio-temporal decision problem~\cite{wang2023time}, where platforms must determine which orders to serve and how to assign heterogeneous couriers at each decision epoch. 
%
%
To address the complexity of concurrent dispatch, recent work has applied reinforcement learning (RL) to concurrent dispatch, modeling couriers or courier-order pairs as learning agents parameterized by spatial and temporal features~\cite{wang2023time, jiang2023faircod, guo2021concurrent}. While these methods demonstrate strong performance in controlled settings, they typically rely on the assumption of full observability of courier states and exclusive platform control. Such  assumptions are reasonable in single-platform systems but no longer hold in realistic multi-platform environments, where couriers may serve multiple platforms simultaneously and platform visibility is inherently limited. Existing work has also studied courier acceptance behavior~\cite{ausseil2024online} and capacity management~\cite{savelsbergh2022challenges} in multi-platform delivery systems. However, these works do not explicitly address how a platform should make dispatch decisions under competition and partial observability. Only a small number of studies investigate courier dispatch under partial observation. These works primarily focus on data privacy-preserving mechanism design~\cite{yang2022privacy, cheng2020real} to facilitate multi-platform courier participation. Nevertheless, they still assume exclusive courier affiliation with a single platform or exclusive control, limiting their applicability to decentralized multi-platform settings. 
To address these challenges, we propose \textbf{POLO}, a \textbf{P}artially \textbf{O}bservable multi-agent reinforcement \textbf{L}earning framework for multi-platform dispatch \textbf{O}ptimization. 
POLO explicitly aligns the learning process with the decentralized and privacy-constrained nature of real-world instant delivery systems. Our main contributions are summarized as follows:  
%
\begin{itemize}
    \item We identify and formalize the instant delivery dispatch problem under partial observability, highlighting the limitation of existing single-platform assumptions. 
  %
  
\item  We propose a partially observable multi-agent RL approach in which each platform-grid pair learns dispatch policies solely from platform-local observations. An attention-based policy representation is introduced to capture heterogeneous courier information under incomplete observations. 

  %
  
\item To mitigate non-stationarity induced by shared couriers and concurrent actions across agents, we design a counterfactual reward shaping mechanism that isolates individual agent contributions. 

  %
  
\item  We develop a high-fidelity simulator based on real-world data and conduct extensive experiments across varying system scales and numbers of platforms, demonstrating consistent improvements in platform revenue and courier travel efficiency over existing methods across varying system scales and numbers of platforms.


\end{itemize}
{
\setlength{\aboverulesep}{0pt}
\setlength{\belowrulesep}{0pt}
\begin{table}[t]
\centering\footnotesize
\caption{Notation Summary}
\label{tab:notation}
\setlength{\tabcolsep}{-1pt}
\renewcommand{\arraystretch}{1.0}
\setlength{\extrarowheight}{1pt}
\begin{tabular}{
p{0.12\columnwidth}
>{\raggedright\arraybackslash}p{0.42\columnwidth}
p{0.12\columnwidth}
>{\raggedright\arraybackslash}p{0.43\columnwidth}
}
\toprule
\rowcolor{HeaderGray}
\textbf{Symbol} & \textbf{Description} & \textbf{Symbol} & \textbf{Description} \\
\midrule
\multicolumn{2}{l}{\textbf{\textit{\bg{Time and Episode}}}} & \multicolumn{2}{l}{\textbf{\textit{\bg{Courier}}}} \\
$t $ & Continuous time & $\mathcal{C}$ & Set of couriers \\
$t_{d} $ & Discrete timestamp index & $c_j $ & Courier $j$ \\
$q$ & Dispatch round index & $\mathbf{l}_{j}^{t} $ & Courier location\\
$\tau $ & Global decision epoch index & $v_j $ & Courier moving speed\\
$M$ & Total timestamps per episode & $\mathbf{d}_{j}^{t} $ & Movement direction \\
$T $ & Max dispatch rounds & $\mathbf{I}_{j}^{t}$ & Courier income \\
\multicolumn{2}{l}{\textbf{\textit{\bg{Platform}}}} & $\hat{\mathbf{I}}_{j}^{t} $ & Courier average income\\
$N_p $ & Total number of platforms & $\mathcal{T}_{j}^{t}$ & Courier historical trajectory\\
$p$ & Platform index & $\mathcal{R}_{j}^{t}$ & Planned future route of $c_j$ \\
\multicolumn{2}{l}{\textbf{\textit{\bg{Location and Grid}}}} & $\mathcal{R}_{j}^{p,t}$ & Route of $c_j$ observable by platform $p$ \\
$\mathbf{l} $ & A 2D spatial location & $d(\mathcal{R}) $ & Total travel distance of route $\mathcal{R}$ \\
$X_{\max} $ & Boundaries of the city & $|\mathcal{R}| $ & Number of stops in route $\mathcal{R}$ \\
$N $ & Number of grid cells & \multicolumn{2}{l}{\textbf{\textit{\bg{Payment and Revenue}}}} \\
$\mathbf{N}_{c}^{t}$ & Couriers in each cell at time $t$ & $r_p$ & Payment ratio to courier \\
$\mathbf{N}_{o}^{t} $ & Unassigned orders in each cell & $r_c$ & Compensation ratio for overdue \\
$\mathbf{N}_{o}^{p,t} $ & Orders from platform $p$ & $R^{p}$ & Revenue of platform $p$ \\
\multicolumn{2}{l}{\textbf{\textit{\bg{Order}}}} & \multicolumn{2}{l}{\textbf{\textit{\bg{Agent and MDP}}}} \\
$\mathcal{O}$ & Set of all orders in one episode & $n $ & Agent (grid cell) index \\
$o_i $ & Order $i$ & $N_{pr}$ & Pruning size \\
$t_{i,c} $ & Creation time of order $o_i$ & $o_{n}^{p,\tau} $ & Earliest order of agent $n$ at epoch $\tau$\\
$t_{i,pd}$ & Promised delivery time of $o_i$ & $\mathcal{C}_{n}^{p,\tau}$ & Candidate courier set for agent $n$ \\
$v_i $ & Monetary value of order $o_i$ & $\mathcal{S}_{n}^{p,\tau}$ & Observation of agent $n$ \\
$\mathbf{l}_{i,p} $ & Pickup location of order $o_i$ & $a_{n}^{p,\tau} $ & Action of agent $n$ \\
$\mathbf{l}_{i,d} $ & Delivery location of order $o_i$ & $r_{n}^{p,\tau} $ & Reward of agent $n$ \\
$p_i$ & Platform of order $o_i$ & \multicolumn{2}{l}{\textbf{\textit{\bg{Training}}}} \\
$s_{i}^{t} $ & State of order $o_i$ at time $t$ & $\gamma $ & Discount factor \\
$\mathcal{O}^p$ & Orders belonging to platform $p$ & $\alpha_a, \alpha_c$ & Learning rates for actor/critic \\
$\mathcal{O}^{p,\tau}$ & Unassigned orders at epoch $\tau$ & $\delta_{n}^{p,\tau}$ & TD advantage \\
\multicolumn{2}{l}{\textbf{\textit{\bg{Matching Features}}}} & $\rho_f, \rho_d $ & Reward coefficients \\
$dp_{i,j}^{\tau}$ & Pickup distance for $(o_i, c_j)$ & \multicolumn{2}{l}{\textbf{\textit{\bg{Functions and Operators}}}} \\
$dt_{i,j}^{p,\tau}$ & Detour for inserting $o_i$ into $\mathcal{R}_{j}^{p,\tau}$ & $\Pi^{p}(\cdot)$ & Platform-local route projection \\
\multicolumn{2}{l}{\textbf{\textit{\bg{Policy Network}}}} & $\mathbb{I}(\cdot)$ & Indicator function \\
$\pi_{\theta_a}$ & Policy network & $v(\mathcal{R})$ & Value of on-time orders in route \\
$V_{\theta_c}$ & Value network & & \\
\bottomrule
\end{tabular}
\end{table}
}
\section{Problem Definition} 
In this section, we present all relevant definitions, and the notations are summarized in Table~\ref{tab:notation}.
\subsection{Preliminaries} 
\subsubsection{Episode} 
Each episode corresponds to a continuous-time decision process within a specific time period.
\subsubsection{Platform} 
We assume that there are a total of \( N_p \) platforms, indexed by \( p \in \{1, \ldots, N_p\} \). A platform \( p \) is responsible for receiving orders from customers and assigning them to available couriers.
\subsubsection{Location and Grid} 
We use \( \mathbf{l} \) to denote any location in the city, where \( \mathbf{l} = (x, y) \). The spatial boundaries of the city are defined by \( X_{\max} \) and \( Y_{\max} \), such that \( x \in [0, X_{\max}] \) and \( y \in [0, Y_{\max}] \). The city is partitioned into \( N \) disjoint hexagonal grid cells. At time \( t \), we record the number of couriers \( \mathbf{N}_c^{t} \) and the number of unassigned orders \( \mathbf{N}_o^{t} \) in each grid cell. Both \( \mathbf{N}_c^{t} \) and \( \mathbf{N}_o^{t} \) belong to \( \mathbb{Z}_{\geq 0}^{N} \). From the perspective of platform \( p \), the observable unassigned order count \( \mathbf{N}_o^{p,t} \) corresponds to the size of the subset of orders associated with that platform.  Accordingly, we have 
\(
\sum_{p} \mathbf{N}_o^{p,t} = \mathbf{N}_o^{t}.
\)
\subsubsection{Order} 
We use \( \mathcal{O} \) to denote the set of orders that appear in a single episode. Each order \( o_i \in \mathcal{O} \) is represented as a tuple:
\begin{equation}
    o_i = \big(t_{i,c},\, t_{i,pd},\, v_i,\, \mathbf{l}_{i,p},\, \mathbf{l}_{i,d},\, p_i\big),
\end{equation}
here, \( t_{i,c} \) denotes the creation time of order \( o_i \), and \( t_{i,pd} \) denotes its promised delivery time. The variable \( v_i \) represents the order value. The pickup and delivery locations of the order \( o_i \) are denoted by \( \mathbf{l}_{i,p} \) and \( \mathbf{l}_{i,d} \), respectively. The platform to which the order belongs is indicated by \( p_i \in \{1, \ldots, N_p\} \).
Accordingly, we define \( \mathcal{O}^{p} \) as the set of orders belonging to platform \( p \), given by
\(\mathcal{O}^{p} = \{\, o_i \in \mathcal{O} \mid p_i = p \,\}.\)

We assume that the status of an order \( o_i \) at time \( t \), denoted by \( s_i^{t} \), can take one of the following discrete values:
\(s_i^{t} \in \{0, 1, 2, 3, 4,5\},\)
where the states correspond to \emph{released}, \emph{assigned}, \emph{picked up}, \emph{on-time delivered},\emph{ overdue delivered,} and \emph{canceled}, respectively. Here, we assume that an order is canceled by the customer if it is not assigned to any courier by its promised delivery time.

\subsubsection{Courier} 
We use \( \mathcal{C} \) to denote the set of couriers. Each courier \( c_j \in \mathcal{C} \) is initialized as
\(c_j = \big( \mathbf{l}_j^{0},\, v_j \big),\)
where \( \mathbf{l}_j^{0} \) denotes the initial location of courier \( c_j \), and \( v_j \) denotes its moving speed. We flatten the location and direction features and use $s_{j,(1)}^{t}$ and $s_{j,(2)}^{t}$ to represent the scalar features and two-dimensional spatial features of courier $c_j$ at time $t$, respectively, defined as
\begin{equation}
s_{j,(1)}^{t} = \big( 
t,\, 
\mathbf{l}_j^{t},\, 
\mathbf{d}_j^{t},\, 
\mathbf{I}_j^{t},\,
\hat{\mathbf{I}}_j^{t},\,
d(\mathcal{R}_j^{t}),\,
|\mathcal{R}_j^{t}|,\,
\mathbb{I}(|\mathcal{R}_j^{t}|> 0) 
\big)\in \mathbb{R}^{10},
\label{courier_define1}
\end{equation}
\begin{equation}
s_{j,(2)}^{t} = \big(\mathcal{T}_j^{t},\, \mathcal{R}_j^{t}\big)
\in \mathbb{R}^{L_r \times 2}.
\label{courier_define2}
\end{equation}
where \( \mathbf{d}_j^{t} \in \mathbb{R}^{2} \) denotes the moving direction at time \( t \), \(\mathbf{I}_j^{t}\) and \(\hat{\mathbf{I}}_j^{t}\) indicate the courier's income and average income from all couriers. Moreover, \( \mathbb{I}(\cdot) \) denotes the indicator function, which equals 1 if the condition is true and 0 otherwise; here we use it to indicate if the courier is idle or not. \( \mathcal{T}_j^{t} \) represents the movement trajectory of the courier \( c_j \) up to time \( t \), and \( \mathcal{R}_j^{t} \) denotes the planned future route if the courier is currently performing delivery tasks.
\(L_r\) denotes the combined length of the \( \mathcal{T}_j^{t} \) and \( \mathcal{R}_j^{t} \).
\( d(\mathcal{R}_j^{t}),\) and
\(|\mathcal{R}_j^{t}|\) indicate the distance and stops in couriers' routes.
We assume that each courier is eligible to perform tasks from all platforms. Consequently, while the historical trajectory \( \mathcal{T}_j^{t} \) is fully observable to all platforms through the courier's mobile GPS signals, the future route plan \( \mathcal{R}_j^{t} \) remains only partially observable. Let \( \mathcal{R}_j^{p,t} \) denote the route plan observable by platform \( p \), defined as
\begin{equation}
    \mathcal{R}_j^{p,t} \triangleq \Pi^{p}\!\left( \mathcal{R}_j^{t} \right),
\end{equation}
where \( \Pi^{p}(\cdot) \) is a platform-local projection operator that extracts the route segments observable to platform \( p \), whose definition is in Algorithm~\ref{algorithm_projection}
\begin{algorithm}[htbp]
\caption{Platform-local projection operator $\Pi^{p}(\mathcal{R})$}
\label{alg:platform_projection_mask}
\KwIn{Full route plan $\mathcal{R} = \langle \mathbf{l}_1, \mathbf{l}_2, \ldots, \mathbf{l}_{|\mathcal{R}|} \rangle$, platform id $p$}
\KwOut{Partially observable route plan $\mathcal{R}^{p}$}

Initialize $\mathcal{R} \leftarrow [\,]$, $\mathcal{L}^p = \{\mathbf{l}_{i,p}|o_j \in \mathcal{O}^p\}\cup\{\mathbf{l}_{i,d}|o_j \in \mathcal{O}^p\}$ \;
\For{$m \leftarrow 1$ \KwTo $|\mathcal{R}|$}{
    \If{$\mathbf{l}_m \in \mathcal{L}^p$}{        
        Append $\mathbf{l}_m$ to $\mathcal{R}^{p}$\;
    }
}
\Return{$\mathcal{R}^{p}$}\;
\label{algorithm_projection}
\end{algorithm}



\begin{figure}     
\centering     
\includegraphics[width=1.0\linewidth]{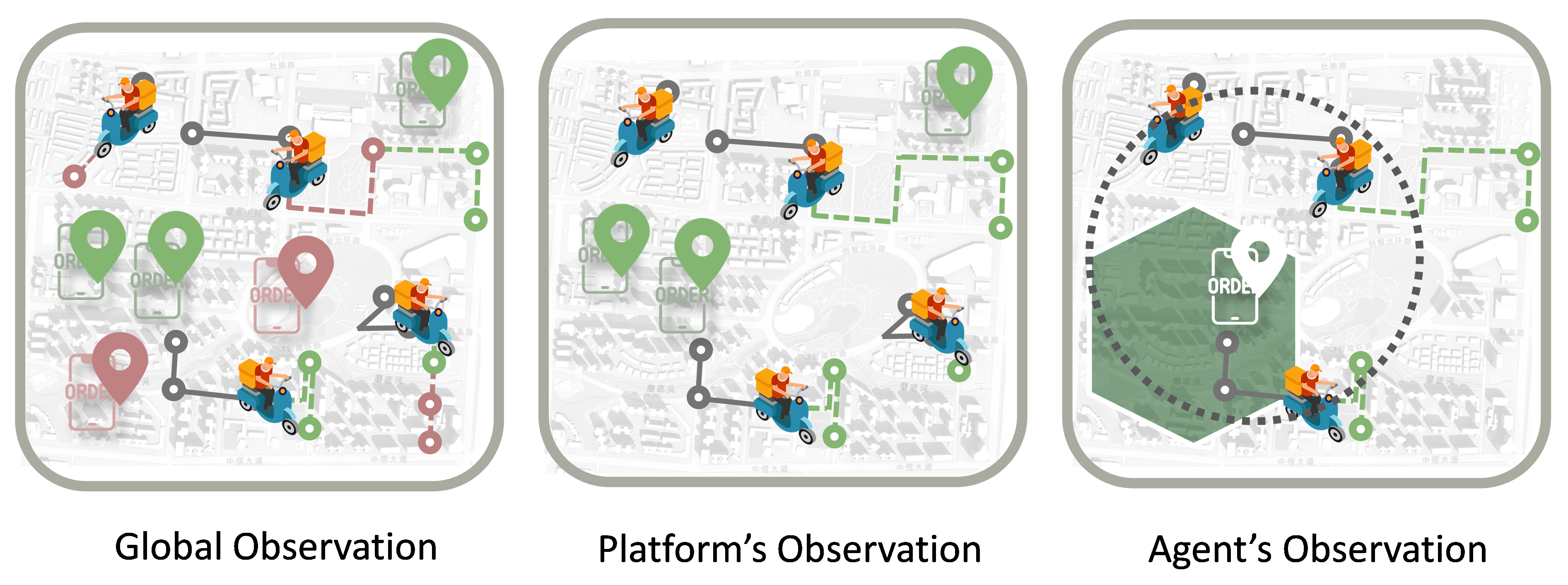}     
\caption{Global state $\rightarrow$ Platform's observation $\rightarrow$ Agent's observation. Each platform observes only its own orders and a platform-local partial route for each courier. The platform agent observes the earliest order and candidate couriers.
}   
\descriptionlabel{}
\label{fig:observation_illustration} 
\vspace{-10pt}
\end{figure} 
\subsubsection{Decision Epoch} 
We divide each episode into \( M \) discrete timestamps, which is a common setting in existing literature and real-world applications ~\cite{jiang2023faircod,wang2025coopride}. Accordingly, the timestamp is indexed by \( t_d \in \{1, 2, \ldots, M\} \). We further assume that at each timestamp \( t_d \), there are at most \( Q_{t_d} \) rounds to sequentially dispatch unassigned orders. For notational simplicity, we linearize all dispatch rounds into a single global decision epoch index \( \tau = \tau(t_d, q) \), where \( q \) denotes the \( q \)-th dispatch round at real timestamp \( t_d \). The global epoch index satisfies \( \tau \in \{0, 1, \ldots, T\} \), where \( T \) denotes the maximum accumulated number of dispatch rounds within an episode.

\subsubsection{Dispatch} 
At each decision epoch \( \tau \), each platform aims to make dispatch decisions by assigning its currently unassigned orders to available couriers. Specifically, the set of unassigned orders for platform \( p \) at decision epoch \( \tau \) is 
\begin{equation}
    \mathcal{O}^{p,\tau} = \{\, o_i \in \mathcal{O}^{p} \mid s_i^{\tau} = 0 \,\}.
\end{equation}

\subsubsection{Payment, Compensation and Revenue}
We assume that a fixed ratio \( r_p \) of the order value is paid to the courier as payment if the courier accepts the order. We have \( s_i^{T} \in \{3, 4, 5\} \) denoted the terminal state of order \( o_i \). If an order is delivered but overdue, an additional compensation with a ratio \( r_c \) is paid to the customer. 
Thus, the revenue of platform \( p \) is defined as follows:
\begin{equation}
    R^{p}
=
\sum_{o_i \in \mathcal{O}^{p}}
\mathbb{I}[s_i^{T} \neq 5]
\Big( (1 - r_p) - r_c \, \mathbb{I}[s_i^{T} = 4] \Big)\, v_i.
\end{equation}

\subsubsection{Decision Objective} We aim to design a dispatch algorithm that makes decisions based on platform-local partial observations, with the goal of maximizing overall platform revenue while simultaneously minimizing the average courier travel distance.

\begin{figure*}
    \centering
    \includegraphics[width=0.8\linewidth]{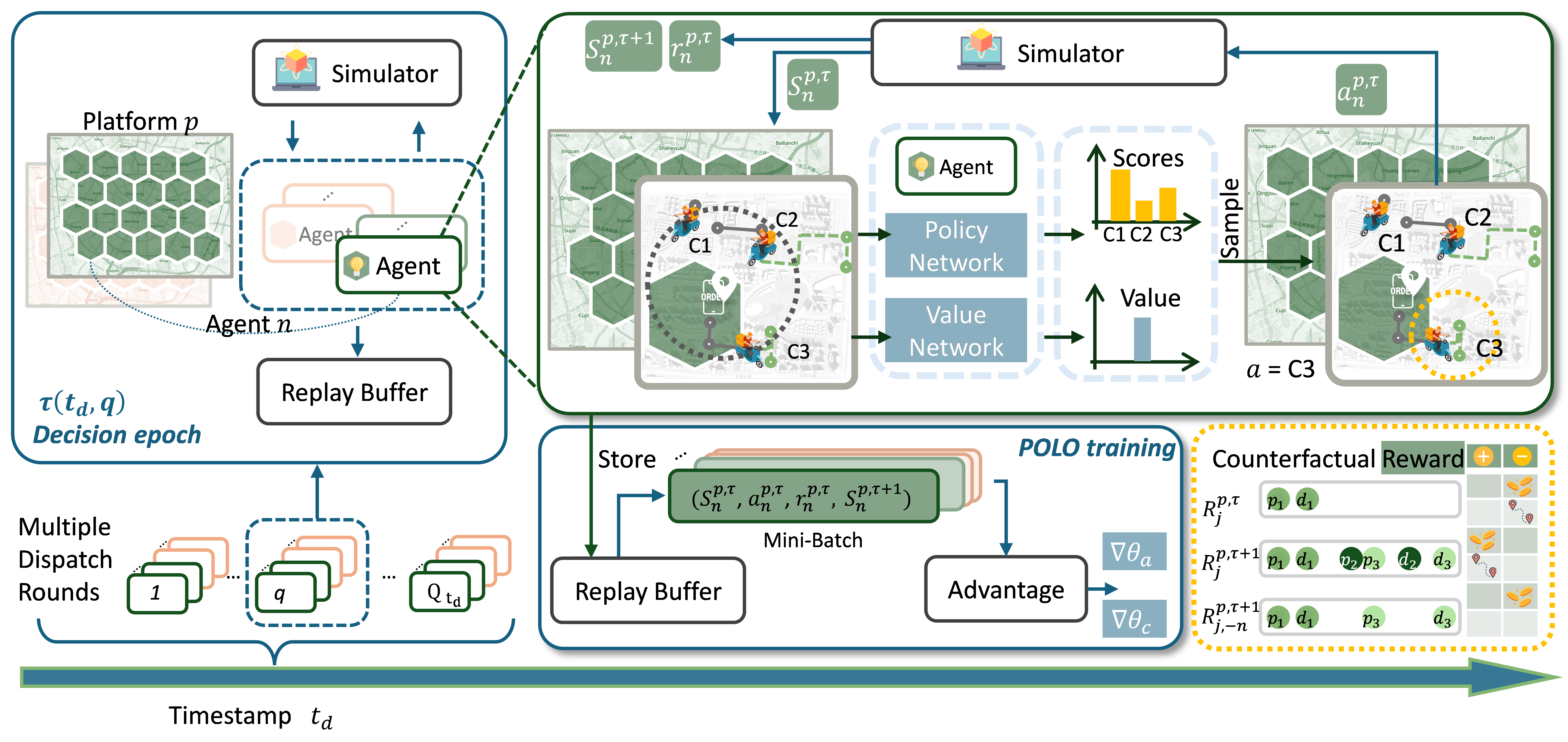}
    \caption{Overview of the POLO framework. The entire decision horizon is divided into discrete timestamps. At each timestamp, agents perform multiple dispatch rounds to make assignment decisions. Each agent adopts an actor–critic architecture, which outputs a score for each candidate courier as well as a value estimate of the current state.}
    \label{fig:Overview}
    \vspace{-8pt}
\end{figure*}
\subsection{Problem Formulation} 

\subsubsection{Platform's Partial Observation} Based on the above definitions, we replace the continuous time index \( t \) with the global decision epoch index \( \tau \) to represent the observed state at decision epoch \( \tau \). As shown in Fig. ~\ref{fig:observation_illustration}, the observation of platform \( p \) consists of four components. The first component is the courier state set $\mathcal{S}_{j}^{p,\tau}$, where each element corresponds to the platform-observable state $s_{j}^{p,\tau}$ of courier $c_j$.
It is constructed by applying platform-local filtering to the states defined in
Eqs.~\ref{courier_define1} and \ref{courier_define2}, as follows:
\begin{equation}
    s_{j}^{p,\tau} = (s_{j,(1)}^{p,\tau},s_{j,(2)}^{p,\tau}).
    \label{courier_state}
\end{equation}


The second observation component, denoted by \( \mathcal{S}_{i}^{p,\tau} \), represents the set of unassigned orders belonging to platform \( p \) at decision epoch \( \tau \). This component consists of \( \lvert \mathcal{O}^{p,\tau} \rvert \) unassigned orders. For each order \( o_i \), we have
\begin{equation}
    s_{i}^{p, \tau} = \bigl( \mathbf{l}_{i,p},\; \mathbf{l}_{i,d},\; v_{i},\; t_{i,pd}-t_d\bigr). 
    \label{order_state}
\end{equation}

We further construct an order-courier matching feature \( \mathcal{S}_{i,j}^{p,\tau} \), which captures features for each feasible order-courier pair. Each element is defined as 
\begin{equation}
    s_{i,j}^{p,\tau} = \bigl( dp_{i,j}^{\tau},\; dt_{i,j}^{^{p,\tau}} \bigr), 
    \label{match_state}
\end{equation}

where \( dp_{i,j}^{\tau} \) denotes the distance between the current location of courier \( c_j \) and the pickup location of order \( o_i \). The term \( dt_{i,j}^{p,\tau} \) represents the incremental travel distance incurred by inserting order \( o_i \) into courier \( c_j \)'s route based on platform \( p \)'s observable route \( \mathcal{R}_{j}^{p,\tau} \).

The fourth state component captures the demand–supply information under partial observability, defined as
\begin{equation}
    \mathcal{S}_{co}^{p,\tau} = \bigl( \mathbf{N}_{c}^{\tau},\; \mathbf{N}_{o}^{p,\tau} \bigr). 
    \label{co_state}
\end{equation}
\subsubsection{Platforms' Spatial Agents} 
We formulate the dispatch problem within a multi-agent reinforcement learning  (MARL) framework. Specifically, we treat each grid cell of each platform as an agent, resulting in a total of \( N \times N_p \) agents. Unassigned orders are spatially partitioned among agents according to their pickup locations. At multiple dispatch rounds, orders assigned to each agent are sorted according to their release times and processed sequentially. Let \( o_{n}^{p,\tau} \) denote the earliest order handled by the \( n \)-th agent with platform \( p \) at decision epoch \( \tau \). For each selected order, we prune the candidate courier set based on the pickup distance and obtain a candidate courier set \( \mathcal{C}_{n}^{p,\tau} \) with size \( N_{pr} \). We then define the four key components of a Markov decision process (MDP) for each agent, namely the observation, action, reward, and state transition, to model the dispatch decision-making process.

\subsubsection{Observation} 
\label{agents_observation}
We use $\mathcal{S}_{n}^{p,\tau}$ to denote the observation of the $n$-th agent. 
It is obtained by filtering the states defined in 
Eqs.~\ref{courier_state}, \ref{order_state}, \ref{match_state}, and \ref{co_state}
with respect to the earliest order $o_{n}^{p,\tau}$, the pruned candidate courier set
$\mathcal{C}_{n}^{p,\tau}$, and the demand–supply state $\mathcal{S}_{co}^{p,\tau}$, as follows:
\begin{equation}
 \mathcal{S}_{n}^{p,\tau} =
 \bigl(
   \mathcal{S}_{j,n}^{p,\tau},
   s_{i,n}^{p,\tau},
   \mathcal{S}_{i,j,n}^{p,\tau},
   \mathcal{S}_{co}^{p,\tau}
 \bigr).
 \label{state}
\end{equation}
Here,
$\mathcal{S}_{j,n}^{p,\tau} = \bigl(\mathcal{S}_{j,(1),n}^{p,\tau}, \mathcal{S}_{j,(2),n}^{p,\tau}\bigr)
\subseteq \mathcal{S}_{j}^{p,\tau}$ denotes the states of the pruned candidate couriers,
where $c_j \in \mathcal{C}_{n}^{p,\tau}$.
The term $s_{i,n}^{p,\tau} \subseteq \mathcal{S}_{i}^{p,\tau}$ corresponds to the earliest
order, where $o_i = o_{n}^{p,\tau}$.
$\mathcal{S}_{i,j,n}^{p,\tau} \subseteq \mathcal{S}_{i,j}^{p,\tau}$ represents
all order-courier matching states between the earliest order
$o_{n}^{p,\tau}$ and each candidate courier $c_j \in \mathcal{C}_{n}^{p,\tau}$.

\subsubsection{Action} Each agent takes its observation as input and outputs a score vector that represents the dispatch preference over candidate couriers. Based on this score vector, an action is defined as
\begin{equation}
    a_{n}^{p,\tau} \in \{0, 1, \ldots, N_{pr}-1\}.
\end{equation} 
The action indicates the index of the selected courier within the candidate courier set  \( \mathcal{C}_{n}^{p,\tau} \). The selected courier is then assigned to serve the order \( o_{n}^{p,\tau} \).

\subsubsection{Reward} 
%
The challenge of reward design arises from two main challenges. First, accepting a new order alters the courier’s route, which may cause previously assigned orders to become overdue. Second, multiple agents may simultaneously influence the same courier in a single dispatch round. 
%
To address these issues, we adopt a counterfactual reward shaping to evaluate the contribution of each agent’s dispatch decision. Let \( \mathcal{R}_{j}^{p,\tau+1} \) denote the updated route after one dispatch round and \( \mathcal{R}_{j,-n}^{p,\tau+1} \) denote the counterfactual route obtained by removing the dispatch action of agent \( n \) from the joint decision.
%
%
We have two functions, \( d(\cdot) \) and \( v(\cdot) \), which compute the total travel distance of a route and the total value of on-time delivered orders along the route, respectively. Based on these definitions, the reward for agent \( n \) taking the dispatch action \( (o_i \rightarrow c_j) \) is computed as
\begin{equation}
\begin{aligned}
r_n^{p,\tau}
= {} & v(\mathcal{R}_{j}^{p,\tau+1})-v(\mathcal{R}_{j}^{p,\tau})
     + \rho_f [v(\mathcal{R}_{j}^{p,\tau+1})-v(\mathcal{R}_{j,-n}^{p,\tau+1})] \\
     & + \rho_d [d(\mathcal{R}_{j}^{p,\tau+1})-d(\mathcal{R}_{j}^{p,\tau})].
\end{aligned}
\label{reward}
\end{equation}

\subsubsection{State Transition} 
After each agent executes a dispatch action, we apply the courier behavior model to determine whether the courier accepts the assigned order. If the courier accepts, the order is inserted into the courier's current route using a detour-minimax insertion heuristic. The system then transitions to decision epoch $\tau \leftarrow\tau+1$.
If there remain unprocessed orders at the current timestamp, the physical time does not advance and the index $t_d$ remains unchanged. Once all orders at the current timestamp have been processed, the system advances the physical time by one decision interval, i.e., $t_d \leftarrow t_d + 1$, and constructs the next state using newly arrived orders in the subsequent interval.

\section{Methodology}
Fig.~\ref{fig:Overview} shows the overview of the proposed MARL framework. The entire decision horizon is divided into discrete timestamps. At each timestamp, agents perform multiple dispatch rounds to make assignment decisions. Each agent takes agent-specific, platform-local observations (as illustrated in Section~\ref{agents_observation}) as input and outputs a score over candidate couriers for dispatch.
\subsection{Policy Network}
The policy network consists of three main components. First, an \emph{intra-courier encoder} is used to encode the individual features of each candidate courier. Next, an \emph{inter-courier encoder} aggregates information across the entire candidate courier set to selectively capture interactions and relative comparisons among candidates. Finally, a multilayer perceptron (MLP) is applied to produce the final output with the desired dimensionality.

\begin{figure}
    \centering
    \includegraphics[width=0.9\linewidth]{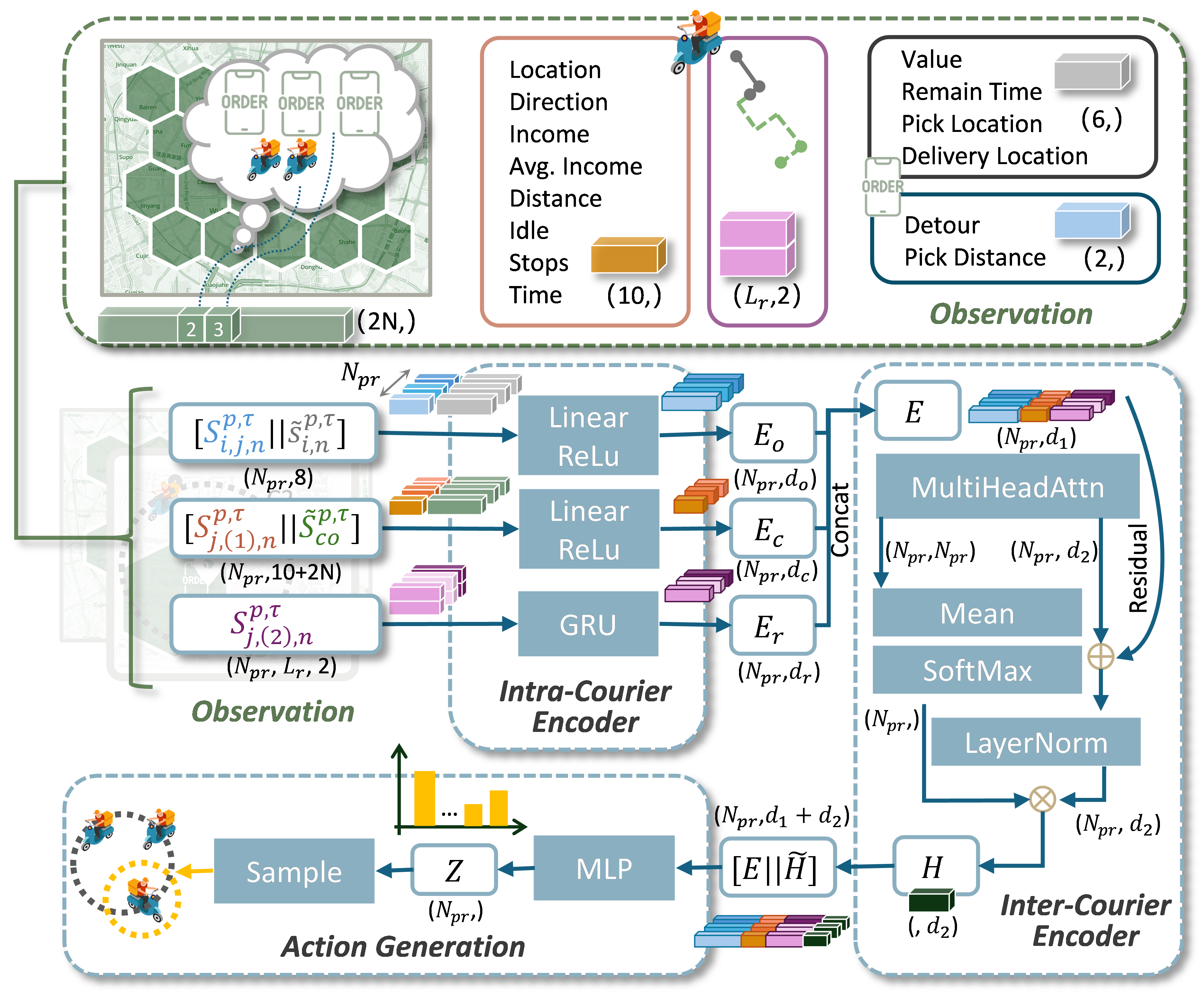}
    \caption{Framework of the policy network. The policy network operates on partial observations of couriers, orders, courier–order matches, and supply–demand states (Eq.~\ref{state}). It applies an intra-encoder to individual features and an inter-encoder to obtain aggregated representations, which are then fed into an MLP to produce the final scores.}
    \vspace{-10pt}
    \label{fig:policy_network}
\end{figure}

\subsubsection{Intra-Courier Encoder}
The policy network takes each agent’s observation \( \mathcal{S}_{n}^{p,\tau} \) (Eq.~\ref{state}) as input and outputs an action index \( a_{n}^{p,\tau} \). As illustrated in Fig.~\ref{fig:policy_network}, the supply–demand feature \( \mathcal{S}_{co}^{p,\tau} \) and the order feature \( s_{i,n}^{p,\tau} \) are repeated \( N_{pr} \) times and concatenated with the candidate couriers’ first feature \( \mathcal{S}_{j,(1),n}^{p,\tau} \) and the courier–order matching feature \( \mathcal{S}_{i,j,n}^{p,\tau} \).
Specifically, we construct two feature vectors
\(
\bigl[ \mathcal{S}_{j,(1),n}^{p,\tau} \,\|\, \tilde{\mathcal{S}}_{co}^{p,\tau} \bigr]
\) and \(
\bigl[ \mathcal{S}_{i,j,n}^{p,\tau} \,\|\, \tilde{s}_{i,n}^{p,\tau} \bigr],
\)
which represent the comprehensive courier features and order-related features, respectively. These concatenated features are then processed by two separate linear encoders to produce the courier and order embeddings
\( \mathbf{E}_c \in \mathbb{R}^{N_{pr} \times d_c} \) and
\( \mathbf{E}_o \in \mathbb{R}^{N_{pr} \times d_o} \).
For the trajectory and route feature \( \mathcal{S}_{j,(2),n}^{p,\tau} \), we employ a gated recurrent unit (GRU) encoder to obtain its embedding representation
\( \mathbf{E}_r \in \mathbb{R}^{N_{pr} \times d_r} \).

\subsubsection{Attention-based Inter-Courier Aggregation Encoder}
Rather than independently evaluating the matching score for each candidate courier, we introduce a novel attention-based mechanism into the policy network to jointly assess all candidates by explicitly considering the entire candidate courier set. Specifically, this module operates as follows. We first concatenate the encoded courier, order, and route embeddings to construct an individual embedding for each candidate courier, given by:
\begin{equation}
\mathbf{E}
=
[\mathbf{E}_c||\mathbf{E}_o||\mathbf{E}_r]
\in
\mathbb{R}^{N_{pr} \times d_1},
\end{equation}
where \(d_1 = d_c + d_o + d_r\). Then we apply a
multi-head self-attention mechanism over the concatenated embeddings
\( \mathbf{E} \) to obtain an aggregated representation \( \mathbf{H} \) as follows: 
\begin{equation}
\mathbf{H}
=
\mathrm{MultiHeadAttn}(\mathbf{E})
\in
\mathbb{R}^{d_2}.
\end{equation}
The attention mechanism enables the policy to model implicit interactions and
competition among different couriers when making effective dispatch decisions.

\subsubsection{Action Generation}
We further repeat the aggregated representation \( \mathbf{H} \) and concatenate it with the individual embeddings \( \mathbf{E} \) to obtain
\( [\tilde{\mathbf{H}} \,\|\, \mathbf{E}] \in
\mathbb{R}^{N_{pr} \times (d_1+d_2)}\).
The concatenated representations are then fed into an MLP to produce
per-courier matching scores:
\begin{equation}
\mathbf{Z}
=
\mathrm{MLP}\!\left(
\big[\tilde{\mathbf{H}} \,\|\, \mathbf{E}\big]
\right)
\in
\mathbb{R}^{N_{pr}}.
\end{equation}
Finally, we define the categorical distribution over actions as:
\begin{equation}
\pi_{\theta_a}(a \mid \mathcal{S})
=
\mathrm{Softmax}(\mathbf{Z}).
\end{equation}
Actions are sampled from this distribution as:
\begin{equation}
a_{n}^{p,\tau}
\sim
\pi_{\theta_a}\!\left(\cdot \mid \mathcal{S}_{n}^{p,\tau}\right).
\end{equation}

\begin{figure*}
    \centering
    \includegraphics[width=0.8\linewidth]{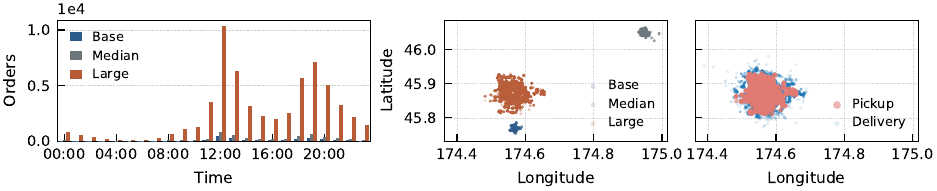}
    \caption{Spatio-temporal distribution of generated orders shows two daily peaks. The delivery area is more spatially extensive than the pickup area.}
    \label{fig:order_distribution}
    \vspace{-8pt}
\end{figure*}
\subsection{Training}
We train the policy network with parameters \( \theta_a \) using an actor-critic framework. The critic network, parameterized by \( \theta_c \), takes the same input \( \mathcal{S}_{n}^{p,\tau} \) as the policy network and is used to estimate the state value function
\( V_{\theta_c}(\mathcal{S}_{n}^{p,\tau}) \), which reflects the expected discounted long-term revenue.
The value function for platform \( p \) at decision epoch \( \tau \) is defined as:
\begin{equation}
V_{\theta_c}\!\left(\mathcal{S}_{n}^{p,\tau}\right)
=
\mathbb{E}
\left[
\sum_{\ell=0}^{T-\tau}
\gamma^{\ell}
\, r_{n}^{p,\tau+\ell}
\right],
\end{equation}
where \( \gamma \in (0,1) \) is the discount factor.
During training, we store transitions from each agent of each platform in a replay buffer $\mathcal{D}$ in the form
\( (\mathcal{S}_{n}^{p,\tau}, a_{n}^{p,\tau}, \mathcal{S}_{n}^{p,\tau+1}, r_{n}^{p,\tau}) \).
At each update step, a mini-batch \( \mathcal{B} \) of transitions is uniformly sampled from $\mathcal{D}$.
We define the temporal-difference (TD) advantage for each transition as:
\begin{equation}
    \delta_{n}^{p,\tau}
=
r_{n}^{p,\tau}
+
\gamma
V_{\theta_c}\!\left(\mathcal{S}_{n}^{p,\tau+1}\right)
-
V_{\theta_c}\!\left(\mathcal{S}_{n}^{p,\tau}\right).
\label{advantage}
\end{equation}

The critic parameters are updated by minimizing the TD error over the mini-batch:
\begin{equation}
    \theta_c
\leftarrow
\theta_c
+
\alpha_c
\sum_{(\mathcal{S}_{n}^{p,\tau},\cdot)\in\mathcal{B}}
\delta_{n}^{p,\tau}
\nabla_{\theta_c}
V_{\theta_c}\!\left(\mathcal{S}_{n}^{p,\tau}\right), 
\label{c_update}
\end{equation}
here \( \alpha_c \) is the learning rate of the critic.
Similarly, the policy parameters are updated with the learning rate  \( \alpha_a \) using the policy gradient:
\begin{equation}
    \theta_a
\leftarrow
\theta_a
+
\alpha_a
\sum_{(\mathcal{S}_{n}^{p,\tau},\cdot)\in\mathcal{B}}
\delta_{n}^{p,\tau}
\nabla_{\theta_a}
\log
\pi_{\theta_a}
\!\left(
a_{n}^{p,\tau}
\mid
\mathcal{S}_{n}^{p,\tau}
\right).
\label{a_update}
\end{equation}
{
\setlength{\aboverulesep}{0pt}
\setlength{\belowrulesep}{0pt}
\begin{table}[t]
\centering
\caption{Statistics of the instances across the three scales.}
\label{tab:agent-dataset-stats}
\small
\setlength{\tabcolsep}{5pt}
\renewcommand{\arraystretch}{1.08}
\begin{tabular}{lccccc}
\toprule
\rowcolor{HeaderGray}
Scale & Orders & Couriers & Time span & Hex size & Grids $N$ \\
\midrule
Base & 86 & 26 & 17:00-18:00 & 0.7 km & 16 \\
Median & 201 & 52 & 17:00-18:00 & 0.7 km & 20 \\
Large & 1418 & 329 & 17:30-18:00 & 0.7 km & 120 \\
\bottomrule
\end{tabular}
\end{table}
}
\begin{algorithm}[htbp]
\SetAlgoNlRelativeSize{-1}
\SetAlgoInsideSkip{smallskip}   
\caption{POLO Training}
\label{alg:polo}

\KwIn{Order set $\mathcal{O}$, courier set $\mathcal{C}$, platforms number $N_p$}
\KwOut{Learned policy $\pi_{\theta_a}$}

Initialize policy network $\pi_{\theta_a}$ and value network $V_{\theta_c}$\;
Initialize replay buffer $\mathcal{D}$\;

\For{episode $= 1$ \KwTo $\texttt{maxEpisode}$}{
    Reset simulator with randomly generated instance\;

    \For{decision epoch $\tau = 0$ \KwTo $T$}{
        \For{each platform $p$ \textbf{in parallel}}{
            Partition $\mathcal{O}^{p,\tau}$ to agents by pickup grid cell\;

            \For{each agent $n$ with order $o_{n}^{p,\tau}$}{
                1. Prune courier pool to obtain $\mathcal{C}_{n}^{p,\tau}$\;
                2. Construct $\mathcal{S}_{n}^{p,\tau}$ according to Eq.~\ref{state}\;
                3. Sample $a_{n}^{p,\tau} \sim \pi_{\theta_a}(\cdot \mid \mathcal{S}_{n}^{p,\tau})$\;
                4. Assign $o_{n}^{p,\tau}$ to courier $c_{a_{n}^{p,\tau}}$\;
            }
        }
        Execute actions and advance to $\tau + 1$\;

        \For{each agent $n$ of platform $p$}{
            1. Compute reward $r_{n}^{p,\tau}$ according to Eq.~\ref{reward}\;
            2. Store $(\mathcal{S}_{n}^{p,\tau}, a_{n}^{p,\tau}, r_{n}^{p,\tau}, \mathcal{S}_{n}^{p,\tau+1})$ in $\mathcal{D}$\;
        }
    }
    \For{ update times $= 1$ \KwTo $\texttt{maxUpdate}$}{
    1. Sample mini-batch $\mathcal{B}$ from $\mathcal{D}$\;
    2. Compute TD advantage according to Eq.~\ref{advantage}\;
    3. Update $\theta_c$ according to Eq.~\ref{c_update}\;
    4. Update $\theta_a$ according to Eq.~\ref{a_update}\;}
}
\end{algorithm}

\section{Evaluation}

\subsection{Simulator}
We construct a simulator based on a real-world dataset collected from a major food delivery platform in China, Meituan\footnote{\url{https://www.meituan.com}}. The dataset contains detailed order information generated over eight consecutive days.
To simulate the demand pattern, in each episode, the simulator randomly samples orders from the same time interval across the eight days to ensure real pickup and delivery location distribution. In the multi-platform scenario, sampled orders are further randomly assigned to different platforms. 
We divide the spatially separated regions into three instance scales, namely Base, Median, and Large. 
%
We visualize the spatio-temporal distribution of the generated orders in Fig.~\ref{fig:order_distribution}. The order demand exhibits two pronounced peaks during daytime hours. Spatially, pickup locations are more concentrated in the city center, while delivery locations are more dispersed toward the surrounding areas.

Couriers are initialized using their first observed locations of the day in the real data. Considering the flexibility of courier working schedules, we randomly activate \( r_a\% \) couriers, which is consistent with empirical findings reported in~\cite{wang2025meituan}, where approximately 67\% of couriers perform the majority of instant delivery tasks.
We also adopt the courier behavior model proposed in~\cite{wang2025meituan}. In this model, a courier’s order rejection probability depends on the number of on-hand orders, the pickup distance, and the remaining time of the most urgent order in the courier’s route.

\subsection{Experiment Setting}
%
{
\setlength{\aboverulesep}{0pt}
\setlength{\belowrulesep}{0pt}
\begin{table*}[t]
\centering
\caption{Overall performance across the three instance scales under different platform counts. Higher GMV and lower ACTD are better. Dark red bold and underline denote the best and second-best results in each column. POLO is lightly shaded, and the last row in the GMV block reports POLO's relative gain over the second-best method.}
\label{tab:overall-reader-acm}
\small
\setlength{\tabcolsep}{4.0pt}
\renewcommand{\arraystretch}{1.10}
\resizebox{\textwidth}{!}{%
\begin{tabular}{llccccccccc}
\toprule
\rowcolor{HeaderGray}
 &  & \multicolumn{3}{c}{\textbf{Base}} & \multicolumn{3}{c}{\textbf{Median}} & \multicolumn{3}{c}{\textbf{Large}} \\
\cmidrule(lr){3-5}\cmidrule(lr){6-8}\cmidrule(lr){9-11}
Metric & Method & $N_p=1$ & $N_p=2$ & $N_p=3$ & $N_p=1$ & $N_p=2$ & $N_p=3$ & $N_p=1$ & $N_p=2$ & $N_p=3$ \\
\midrule
\multirow{9}{*}{\textbf{GMV} ($\uparrow$)} & Random & $158.58_{\pm 14.21}$ & $160.09_{\pm 12.29}$ & $157.37_{\pm 14.51}$ & $346.92_{\pm 21.22}$ & $353.49_{\pm 25.21}$ & $342.66_{\pm 23.83}$ & $-261.56_{\pm 73.61}$ & $-266.88_{\pm 58.22}$ & $-296.00_{\pm 55.69}$ \\
 & PRandom & $164.04_{\pm 14.72}$ & $166.28_{\pm 12.98}$ & $166.69_{\pm 11.62}$ & $374.64_{\pm 22.28}$ & $374.96_{\pm 27.18}$ & $373.69_{\pm 25.55}$ & $2119.66_{\pm 94.96}$ & $2122.54_{\pm 90.53}$ & $2119.99_{\pm 95.74}$ \\
 & DetGreedy & $133.69_{\pm 12.35}$ & $123.30_{\pm 12.64}$ & $122.04_{\pm 12.18}$ & $326.65_{\pm 23.04}$ & $299.80_{\pm 24.61}$ & $285.16_{\pm 25.92}$ & $1379.78_{\pm 78.93}$ & $1206.06_{\pm 77.10}$ & $1171.64_{\pm 75.75}$ \\
 & DisGreedy & $164.36_{\pm 14.72}$ & $164.86_{\pm 14.83}$ & $163.35_{\pm 14.94}$ & $359.58_{\pm 25.24}$ & $361.33_{\pm 30.83}$ & $363.88_{\pm 27.01}$ & $2207.85_{\pm 90.19}$ & $2194.45_{\pm 94.41}$ & $2212.10_{\pm 94.15}$ \\
 & IBM & $176.51_{\pm 17.13}$ & $159.81_{\pm 16.18}$ & $159.83_{\pm 20.93}$ & $431.68_{\pm 31.20}$ & $374.64_{\pm 30.07}$ & $367.63_{\pm 32.11}$ & $3847.10_{\pm 121.19}$ & $2848.89_{\pm 100.91}$ & $2582.17_{\pm 115.46}$ \\
 & TCAC & \textcolor{red!80!black}{$\bm{220.94}_{\bm{\pm 7.65}}$} & \textcolor{red!80!black}{$\bm{207.47}_{\bm{\pm 7.98}}$} & \textcolor{red!80!black}{$\bm{194.42}_{\bm{\pm 10.61}}$} & \underline{$512.56_{\pm 18.41}$} & $464.48_{\pm 21.15}$ & \underline{$443.60_{\pm 20.49}$} & \underline{$4216.44_{\pm 113.94}$} & \underline{$3006.11_{\pm 98.19}$} & \underline{$2875.70_{\pm 107.49}$} \\
 & FAIR & $218.77_{\pm 8.20}$ & $200.35_{\pm 10.70}$ & $178.36_{\pm 12.36}$ & $494.84_{\pm 20.72}$ & \underline{$477.54_{\pm 20.41}$} & $439.18_{\pm 20.62}$ & $2973.70_{\pm 102.78}$ & $2964.76_{\pm 92.24}$ & $2745.94_{\pm 97.85}$ \\
\rowcolor{green!8}  & \textbf{POLO} & \underline{$219.56_{\pm 8.56}$} & \underline{$202.81_{\pm 9.34}$} & \underline{$183.82_{\pm 10.28}$} & \textcolor{red!80!black}{$\bm{516.43}_{\bm{\pm 18.13}}$} & \textcolor{red!80!black}{$\bm{481.99}_{\bm{\pm 20.68}}$} & \textcolor{red!80!black}{$\bm{455.92}_{\bm{\pm 18.14}}$} & \textcolor{red!80!black}{$\bm{4229.25}_{\bm{\pm 135.20}}$} & \textcolor{red!80!black}{$\bm{3290.76}_{\bm{\pm 100.20}}$} & \textcolor{red!80!black}{$\bm{2911.20}_{\bm{\pm 107.14}}$} \\
\rowcolor{green!8}  & \textbf{POLO Gain (\%)} & $-0.62\%$ & $-2.24\%$ & $-5.45\%$ & $+0.76\%$ & $+0.93\%$ & $+2.78\%$ & $+0.30\%$ & $+9.47\%$ & $+1.23\%$ \\
\midrule
\multirow{8}{*}{\textbf{ACTD} ($\downarrow$)} & Random & $6.28_{\pm 0.37}$ & $6.37_{\pm 0.34}$ & $6.37_{\pm 0.33}$ & $8.18_{\pm 0.28}$ & $8.10_{\pm 0.29}$ & $8.15_{\pm 0.25}$ & $20.79_{\pm 0.31}$ & $20.86_{\pm 0.24}$ & $20.88_{\pm 0.29}$ \\
 & PRandom & $5.17_{\pm 0.29}$ & $5.19_{\pm 0.24}$ & $5.19_{\pm 0.28}$ & $6.02_{\pm 0.21}$ & $6.04_{\pm 0.26}$ & $6.05_{\pm 0.25}$ & $10.04_{\pm 0.16}$ & $10.03_{\pm 0.18}$ & $10.04_{\pm 0.20}$ \\
 & DetGreedy & $6.33_{\pm 0.30}$ & $5.81_{\pm 0.29}$ & $5.68_{\pm 0.29}$ & $7.51_{\pm 0.25}$ & $6.72_{\pm 0.28}$ & $6.50_{\pm 0.27}$ & $13.59_{\pm 0.20}$ & $11.98_{\pm 0.20}$ & $11.42_{\pm 0.23}$ \\
 & DisGreedy & \underline{$4.23_{\pm 0.30}$} & \underline{$4.24_{\pm 0.27}$} & \underline{$4.23_{\pm 0.27}$} & \underline{$5.10_{\pm 0.26}$} & \underline{$5.10_{\pm 0.23}$} & \underline{$5.09_{\pm 0.26}$} & $8.68_{\pm 0.16}$ & \underline{$8.69_{\pm 0.17}$} & \underline{$8.71_{\pm 0.18}$} \\
 & IBM & \textcolor{red!80!black}{$\bm{3.82}_{\bm{\pm 0.21}}$} & \textcolor{red!80!black}{$\bm{4.03}_{\bm{\pm 0.25}}$} & \textcolor{red!80!black}{$\bm{4.12}_{\bm{\pm 0.26}}$} & \textcolor{red!80!black}{$\bm{4.47}_{\bm{\pm 0.17}}$} & \textcolor{red!80!black}{$\bm{4.72}_{\bm{\pm 0.18}}$} & \textcolor{red!80!black}{$\bm{4.85}_{\bm{\pm 0.21}}$} & \textcolor{red!80!black}{$\bm{6.33}_{\bm{\pm 0.11}}$} & \textcolor{red!80!black}{$\bm{7.33}_{\bm{\pm 0.15}}$} & \textcolor{red!80!black}{$\bm{7.73}_{\bm{\pm 0.17}}$} \\
 & TCAC & $5.39_{\pm 0.27}$ & $5.40_{\pm 0.27}$ & $5.38_{\pm 0.28}$ & $6.05_{\pm 0.18}$ & $6.18_{\pm 0.21}$ & $6.20_{\pm 0.22}$ & $8.23_{\pm 0.13}$ & $9.63_{\pm 0.17}$ & $10.21_{\pm 0.19}$ \\
 & FAIR & $5.42_{\pm 0.29}$ & $5.46_{\pm 0.28}$ & $5.35_{\pm 0.30}$ & $6.61_{\pm 0.23}$ & $6.32_{\pm 0.23}$ & $6.22_{\pm 0.23}$ & $10.28_{\pm 0.18}$ & $10.28_{\pm 0.16}$ & $10.26_{\pm 0.19}$ \\
\rowcolor{green!8}  & \textbf{POLO} & $5.25_{\pm 0.21}$ & $5.35_{\pm 0.31}$ & $5.29_{\pm 0.29}$ & $6.05_{\pm 0.19}$ & $6.28_{\pm 0.28}$ & $6.15_{\pm 0.25}$ & \underline{$8.08_{\pm 0.11}$} & $9.72_{\pm 0.17}$ & $10.18_{\pm 0.16}$ \\
\bottomrule
\end{tabular}%
}
\end{table*}
}

For the simulator, the decision interval is set to 2 minutes. The compensation ratio is set to \( r_c = 1 \), and the payment ratio is \( r_p = 0.2 \). We randomly activate \(r_a = 50\%\) couriers in each episode for the base- and median-scale settings and \(r_a = 10\%\) for the large-scale setting. The courier travel speed is set to \(v_c = 0.003~km/s\), and the size of the candidate courier set is \(N_{pr} = 10\). For the base- and median-scale settings, each episode spans from 17:00 to 18:00, while for the large-scale setting, each episode spans from 17:30 to 18:00.
For training, the replay buffer size is set to 100,000, and the batch size is 256. The policy network and value network are updated 10 times after each episode, and the models are trained for a total of 1,000 episodes. All agents share network parameters, and both networks are optimized using the Adam optimizer with a learning rate of 0.0002. The discount factor is set to \( \gamma = 0.97 \).
The entire region is partitioned into hexagonal grids with a side length of 0.7~km, and the attention module employs 4 attention heads. Under this setting, we summarize the dataset statistics and the total number of grids in Table~\ref{tab:agent-dataset-stats}.

\begin{figure}
        \centering
        \includegraphics[width=0.75\linewidth]{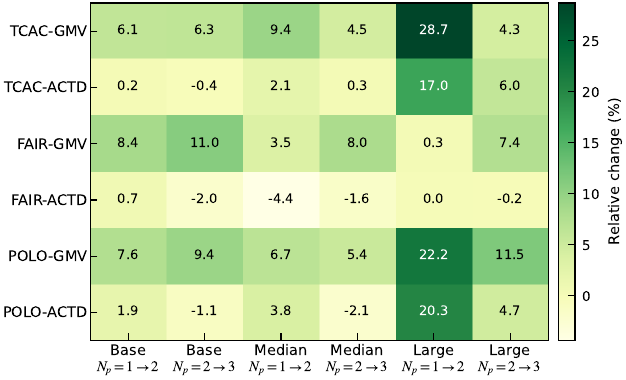}
        \caption{Relative performance change as the number of platforms increases.}
        \label{fig:platform_heatmap}
    \label{fig:combined}
\end{figure}

\subsection{Baseline and Metrics}
We compare our model with the following baselines:
\begin{itemize}[leftmargin=*]
    \item \textbf{Random}: Orders are dispatched randomly to available couriers.
    \item \textbf{PRandom}: Orders are dispatched randomly to the $N_{\text{pr}}$ closest available couriers. This baseline mimics a broadcast-based dispatch mechanism, in which couriers compete to accept orders and the fastest responder is assigned the order.
    \item \textbf{DisGreedy}: Orders are assigned to the closest available courier.
    \item \textbf{DetGreedy}: Orders are assigned to the courier that incurs the minimum additional travel distance.
    \item \textbf{IBM} ~\cite{zheng2018order}: Orders are iteratively assigned using bipartite graph matching to maximize the total matching weight at each decision epoch, where the weight of each order–courier pair is defined based on the order value and the corresponding detour distance.
    \item \textbf{TCAC} ~\cite{wang2023time}: A learning-based concurrent dispatch method that focuses on time constraints.
    \item \textbf{FAIR} ~\cite{jiang2023faircod}: A learning-based concurrent dispatch method that focuses on reducing income disparities among couriers.
\end{itemize}
All baseline methods focus on dispatch decision-making, while the minimax detour strategy is uniformly adopted by the simulator for route planning.
The two learning-based methods, TCAC and FAIR, use the same state construction and training settings as POLO to ensure a fair comparison.
We evaluate the performance of different dispatch methods using three metrics from the \emph{global} perspective:
\textbf{GMV} (Gross Merchandise Volume) and \textbf{ACTD} (Average Courier Travel Distance). The detailed definitions of these metrics are as follows:
\[
GMV = \sum_{p = 1}^{N_p} R^p
, \quad
ACTD = \frac{\sum_{c_j\in \mathcal{C}} d(\mathcal{T}_j^T)}{|\mathcal{C}|}
\]
\begin{figure}
        \centering
        \includegraphics[width=0.9\linewidth]{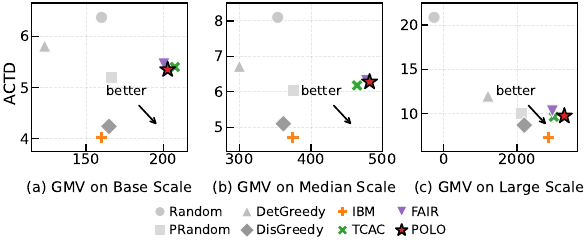}
        \caption{Trade-off analysis between key metrics.}        \label{fig:platform_heatmap}
    \label{fig:scatter}
\end{figure}
\subsection{Overall Performance}

Table~\ref{tab:overall-reader-acm} presents the experimental results under different problem scales and numbers of platforms. We summarize the following findings based on the results.

(1) As the number of platforms increases, all learning-based methods exhibit performance degradation, reflected by reduced GMV, increased ACTD, or both. Moreover, this degradation becomes more significant in large-scale scenarios. For example, when applying the TCAC method in the base-scale setting, increasing the number of platforms from 1 to 2 reduces the GMV from 220.94 to 207.47, corresponding to only a 6.1\% decrease. In contrast, under the large-scale setting, the GMV decreases from 4216.44 to 3006.11, which exceeds a 28.7\% reduction. This finding indicates that partially observable environments can substantially affect decision-making performance in real-world applications, especially in large-scale multi-platform settings.
Among these methods, POLO achieves superior GMV performance in most experimental settings. Although POLO obtains the second-best GMV in the base-scale scenario, its advantage becomes much more significant in large-scale scenarios. For example, under the large-scale setting with two platforms, POLO achieves approximately a 9.47\% improvement in GMV compared to TCAC.

(2) On the other hand, heuristic methods such as IBM and DisGreedy generally achieve lower ACTD values, as they assign orders primarily based on minimum detour distance without considering order value or delivery urgency. While this minimizes courier travel, it sacrifices revenue by ignoring high-value assignments. POLO's reward explicitly balances order value against travel cost through the distance penalty coefficient $\rho_d$, such that the higher ACTD reflects a deliberate design choice prioritizing revenue over travel minimization, rather than a routing inefficiency. Although POLO does not achieve the lowest ACTD compared with heuristic methods, it consistently attains higher GMV and lower ACTD than the other learning-based baselines. To further illustrate the trade-off between these two metrics, Figure~\ref{fig:scatter} presents scatter visualizations under the two-platform large-scale setting. Ideally, methods closer to the bottom-right corner indicate better performance, corresponding to higher GMV and lower ACTD. The results show that POLO consistently occupies a more favorable position across all scales, with particularly clear advantages in large-scale scenarios.

\subsection{Performance under Distribution Shifts}
In real-world applications, different platforms may exhibit different preferences regarding delivery time, pricing strategies, and service quality. To evaluate the robustness of our method under distribution shift scenarios, we further report experimental results under both spatial and temporal distribution shifts. Specifically, for the spatial distribution shift, orders generated from different geographic areas are more likely to belong to different platforms. For the temporal distribution shift, we simulate heterogeneous platform preferences by assigning one platform a lower delivery price but a longer promised delivery time (i.e., lower \(v_i\) and higher \(t_{i,pd}\)).
The corresponding results are presented in Table~\ref{tab:diff-reader-acm}. We observe that our model achieves the best GMV performance in most distribution shift scenarios, demonstrating its strong adaptability and robustness under realistic multi-platform environments.

{
\setlength{\aboverulesep}{0pt}
\setlength{\belowrulesep}{0pt}
\begin{table}[t]
\centering
\caption{Sensitivity results under temporal and spatial distribution shifts with $N_p = 2$. }
\label{tab:diff-reader-acm}
\small
\setlength{\tabcolsep}{2.0pt}
\renewcommand{\arraystretch}{1.10}
\begin{tabular}{llcccc}
\toprule
\rowcolor{HeaderGray}
 & & \multicolumn{2}{c}{\textbf{Temporal diff}} & \multicolumn{2}{c}{\textbf{Spatial diff}} \\
\cmidrule(lr){3-4}\cmidrule(lr){5-6}
Scale & Method & GMV ($\uparrow$) & ACTD ($\downarrow$) & GMV ($\uparrow$) & ACTD ($\downarrow$) \\
\midrule
\multirow{8}{*}{\textbf{Base}} & Random & $152.60_{\pm 9.37}$ & $6.37_{\pm 0.37}$ & $160.07_{\pm 14.99}$ & $6.35_{\pm 0.35}$ \\
 & PRandom & $156.28_{\pm 10.65}$ & $5.24_{\pm 0.31}$ & $166.22_{\pm 12.52}$ & $5.18_{\pm 0.30}$ \\
 & DetGreedy & $124.91_{\pm 12.38}$ & $5.77_{\pm 0.31}$ & $121.44_{\pm 12.28}$ & $5.71_{\pm 0.27}$ \\
 & DisGreedy & $152.72_{\pm 13.29}$ & \underline{$4.26_{\pm 0.25}$} & $166.15_{\pm 15.56}$ & \underline{$4.29_{\pm 0.24}$} \\
 & IBM & $154.12_{\pm 15.51}$ & \textcolor{red!80!black}{$\bm{4.08}_{\bm{\pm 0.27}}$} & $162.76_{\pm 17.28}$ & \textcolor{red!80!black}{$\bm{4.00}_{\bm{\pm 0.23}}$} \\
 & TCAC & $173.03_{\pm 8.78}$ & $5.24_{\pm 0.28}$ & \textcolor{red!80!black}{$\bm{209.15}_{\bm{\pm 9.31}}$} & $5.39_{\pm 0.30}$ \\
 & FAIR & \underline{$175.26_{\pm 7.70}$} & $5.38_{\pm 0.30}$ & $192.71_{\pm 9.30}$ & $5.50_{\pm 0.26}$ \\
\rowcolor{green!8}  & \textbf{POLO} & \textcolor{red!80!black}{$\bm{177.44}_{\bm{\pm 8.44}}$} & $5.30_{\pm 0.27}$ & \underline{$196.52_{\pm 13.05}$} & $5.26_{\pm 0.31}$ \\
\midrule
\multirow{8}{*}{\textbf{Large}} & Random & $158.50_{\pm 57.41}$ & $20.82_{\pm 0.26}$ & $-266.06_{\pm 67.52}$ & $20.82_{\pm 0.27}$ \\
 & PRandom & $2239.22_{\pm 82.25}$ & $10.05_{\pm 0.20}$ & $2106.82_{\pm 98.88}$ & $10.06_{\pm 0.17}$ \\
 & DetGreedy & $1365.81_{\pm 78.59}$ & $11.99_{\pm 0.23}$ & $1195.34_{\pm 74.78}$ & $11.96_{\pm 0.20}$ \\
 & DisGreedy & $2261.62_{\pm 93.19}$ & \underline{$8.64_{\pm 0.18}$} & $2216.57_{\pm 97.18}$ & \underline{$8.66_{\pm 0.18}$} \\
 & IBM & $2884.94_{\pm 110.70}$ & \textcolor{red!80!black}{$\bm{7.31}_{\bm{\pm 0.14}}$} & $2986.11_{\pm 139.91}$ & \textcolor{red!80!black}{$\bm{7.18}_{\bm{\pm 0.16}}$} \\
 & TCAC & \underline{$3180.27_{\pm 89.37}$} & $9.89_{\pm 0.16}$ & \underline{$3154.97_{\pm 109.86}$} & $9.91_{\pm 0.15}$ \\
 & FAIR & $2927.71_{\pm 93.80}$ & $10.29_{\pm 0.18}$ & $2928.44_{\pm 93.63}$ & $10.28_{\pm 0.19}$ \\
\rowcolor{green!8}  & \textbf{POLO} & \textcolor{red!80!black}{$\bm{3258.55}_{\bm{\pm 77.20}}$} & $9.73_{\pm 0.19}$ & \textcolor{red!80!black}{$\bm{3384.15}_{\bm{\pm 119.68}}$} & $9.38_{\pm 0.16}$ \\
\bottomrule
\end{tabular}
\end{table}
}

\subsection{Ablation Study}
{
\setlength{\aboverulesep}{0pt}
\setlength{\belowrulesep}{0pt}
\begin{table}[t]
\centering
\caption{Attention ablation study with $N_p = 2$. }
\label{tab:attention-reader-acm}
\small
\setlength{\tabcolsep}{1.0pt}
\renewcommand{\arraystretch}{1.10}
\begin{tabular}{lcccccc}
\toprule
\rowcolor{HeaderGray}
Variant & \multicolumn{3}{c}{\textbf{Base}} & \multicolumn{3}{c}{\textbf{Large}} \\
\cmidrule(lr){2-4}\cmidrule(lr){5-7}
 & GMV ($\uparrow$) & ACTD ($\downarrow$) & Time ($\downarrow$) & GMV ($\uparrow$) & ACTD ($\downarrow$) & Time ($\downarrow$) \\
\midrule
\rowcolor{green!8}
POLO & \textcolor{red!80!black}{$\bm{202.81}_{\bm{\pm 9.34}}$} & $5.35_{\pm 0.31}$ & \textcolor{red!80!black}{$\bm{0.1601}$} & \textcolor{red!80!black}{$\bm{3290.76}_{\bm{\pm 100.20}}$} & \textcolor{red!80!black}{$\bm{9.72}_{\bm{\pm 0.17}}$} & $4.6703$ \\
w/o Att & $199.22_{\pm 10.68}$ & \textcolor{red!80!black}{$\bm{5.30}_{\bm{\pm 0.31}}$} & $0.1782$ & $3178.27_{\pm 96.43}$ & $9.86_{\pm 0.17}$ & \textcolor{red!80!black}{$\bm{4.2383}$} \\
\bottomrule
\end{tabular}
\end{table}
}
{
\setlength{\aboverulesep}{0pt}
\setlength{\belowrulesep}{0pt}
\begin{table*}[t]
\centering
\caption{Effect of zeroing each reward coefficient in POLO. We compare the zero setting against the reference non-zero setting used in the main configuration ($0.6$ for counterfactual ratio and $1.0$ for distance penalty ratio). We define $\Delta\mathrm{GMV}=\mathrm{GMV}_{\mathrm{ref}}-\mathrm{GMV}_{0}$ and $\Delta\mathrm{ACTD}=\mathrm{ACTD}_{0}-\mathrm{ACTD}_{\mathrm{ref}}$, so positive deltas indicate benefit from using the non-zero coefficient. Positive deltas are highlighted in dark red.}
\label{tab:reward-zero-ablation}
\small
\setlength{\tabcolsep}{2.0pt}
\renewcommand{\arraystretch}{1.10}
\resizebox{\textwidth}{!}{%
\begin{tabular}{lcccccccccccccccc}
\toprule
\rowcolor{HeaderGray}
Coeff. & \multicolumn{8}{c}{\textbf{Base ($N_p = 2$)}} & \multicolumn{8}{c}{\textbf{Large ($N_p = 2$)}} \\
\cmidrule(lr){2-9}\cmidrule(lr){10-17}
 & $0$ & Ref. & GMV@0 & GMV@ref & ACTD@0 & ACTD@ref & $\Delta$GMV & $\Delta$ACTD & $0$ & Ref. & GMV@0 & GMV@ref & ACTD@0 & ACTD@ref & $\Delta$GMV & $\Delta$ACTD \\
\midrule
$\rho_f$ & 0.0 & 0.6 & $196.79_{\pm 11.27}$ & $202.81_{\pm 9.34}$ & $5.21_{\pm 0.28}$ & $5.35_{\pm 0.31}$ & \textcolor{red!80!black}{$\bm{+6.02}$} & $-0.14$ & 0.0 & 0.6 & $3207.47_{\pm 110.31}$ & $3290.76_{\pm 100.20}$ & $9.88_{\pm 0.16}$ & $9.72_{\pm 0.17}$ & \textcolor{red!80!black}{$\bm{+83.29}$} & \textcolor{red!80!black}{$\bm{+0.16}$} \\
$\rho_d $ & 0.0 & 1.0 & $193.08_{\pm 12.61}$ & $202.81_{\pm 9.34}$ & $5.34_{\pm 0.28}$ & $5.35_{\pm 0.31}$ & \textcolor{red!80!black}{$\bm{+9.73}$} & $-0.01$ & 0.0 & 1.0 & $2974.51_{\pm 111.94}$ & $3290.76_{\pm 100.20}$ & $10.10_{\pm 0.20}$ & $9.72_{\pm 0.17}$ & \textcolor{red!80!black}{$\bm{+316.25}$} & \textcolor{red!80!black}{$\bm{+0.38}$} \\
\bottomrule
\end{tabular}%
}
\end{table*}
}

Table~\ref{tab:attention-reader-acm} and Table~\ref{tab:reward-zero-ablation} further analyze the effectiveness of two critical design components in POLO. 
From Table~\ref{tab:attention-reader-acm}, where the variant without the attention module is denoted as \textbf{w/o Att}, we observe that the attention-based inter-courier aggregation mechanism, although introducing additional computational overhead in large-scale scenarios, effectively improves performance on both GMV and ACTD. Moreover, the increase in inference time remains acceptable in practice.
On the other hand, Table~\ref{tab:reward-zero-ablation} evaluates the contributions of different components in the counterfactual reward design. The results indicate that both the counterfactual term and the distance penalty term positively contribute to the overall performance. Furthermore, similar to previous observations, the performance improvements become more significant under large-scale settings, highlighting the potential of POLO in real-world applications.

\subsection{Sensitivity Analysis}
Table~\ref{tab:hex-size-reader-acm} presents the performance of learning-based methods on the large-scale scenario with two platforms under different hexagon sizes. We observe that, although the hexagon size needs to be manually specified, POLO consistently achieves strong performance across different settings, demonstrating its robustness to changes in spatial partition granularity.
\begin{figure*}
    \centering

    \begin{subfigure}[t]{0.7\linewidth}
        \centering
        \includegraphics[width=\linewidth]{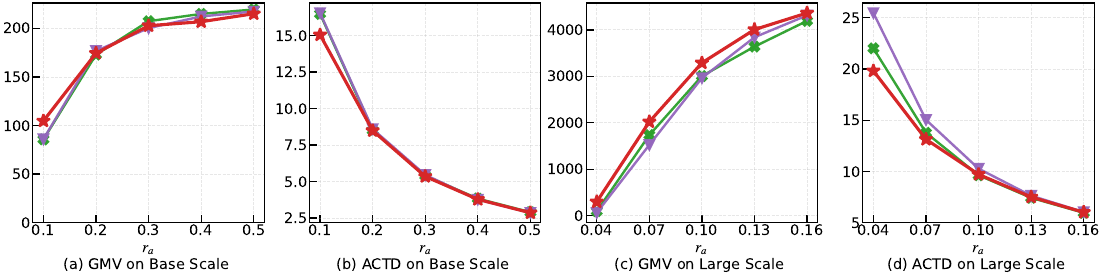}
    \end{subfigure}

    \vspace{0.5em}

    \begin{subfigure}[t]{0.7\linewidth}
        \centering
        \includegraphics[width=\linewidth]{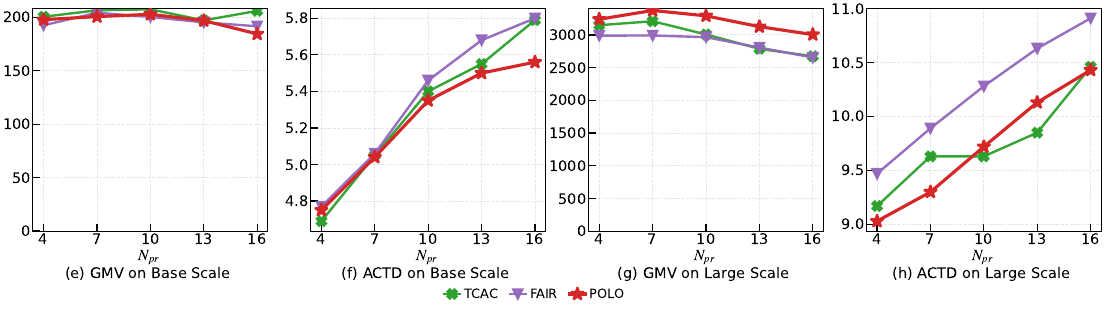}
    \end{subfigure}
    \caption{Sensitivity analysis under different pruning numbers $N_{pr}$ and courier appearance ratios $r_a$. POLO achieves higher GMV and lower ACTD in most parameter settings.}
    \label{courier}
\end{figure*}
Figure~\ref{courier} presents the performance under different pruning numbers $N_{pr}$ and courier appearance ratios $r_a$. Overall, POLO demonstrates robust performance across different parameter settings.
The first parameter, the courier appearance ratio $r_a$, reflects different workload levels in the system. Specifically, it simulates scenarios ranging from limited courier availability to abundant courier resources. As expected, increasing the number of available couriers generally leads to higher GMV and lower ACTD. Notably, even under scenarios with limited courier availability, POLO is still able to achieve relatively high GMV, indicating that the proposed method can identify effective dispatch strategies even in challenging environments.
The second parameter, the pruning number $N_{pr}$, controls the number of candidate couriers considered for each order. The results show that, for POLO, the optimal pruning number is around 10 in terms of GMV performance. When $N_{pr}$ is too small, the candidate set becomes overly restricted, potentially excluding high-quality assignment choices. In contrast, an excessively large $N_{pr}$ increases the difficulty of decision-making and may introduce unnecessary noise into the optimization process.
\subsection{Inference Efficiency}
We report the inference time of all learning-based methods in Table~\ref{tab:large-deep-inference-time}, measured as the average end-to-end seconds per test instance, including feature extraction, decision-making, and simulator stepping. 
We find that POLO achieves the lowest inference time among all methods. This is mainly because TCAC and FAIR make decisions sequentially for each order, while POLO performs parallel order assignment across multiple grids, resulting in faster inference.
{
\setlength{\aboverulesep}{0pt}
\setlength{\belowrulesep}{0pt}
\begin{table*}[t]
\centering
\caption{Hex-size sensitivity on the large instance.}
\label{tab:hex-size-reader-acm}
\small
\setlength{\tabcolsep}{4.0pt}
\renewcommand{\arraystretch}{1.10}
\begin{tabular}{lcccccccc}
\toprule
\rowcolor{HeaderGray}Method & \multicolumn{2}{c}{\textbf{0.4}} & \multicolumn{2}{c}{\textbf{0.7}} & \multicolumn{2}{c}{\textbf{1.0}} & \multicolumn{2}{c}{\textbf{1.3}} \\
\cmidrule(lr){2-3}\cmidrule(lr){4-5}\cmidrule(lr){6-7}\cmidrule(lr){8-9}
 & GMV ($\uparrow$) & ACTD ($\downarrow$) & GMV ($\uparrow$) & ACTD ($\downarrow$) & GMV ($\uparrow$) & ACTD ($\downarrow$) & GMV ($\uparrow$) & ACTD ($\downarrow$) \\
\midrule
TCAC & \underline{$3077.69_{\pm 133.90}$} & \textcolor{red!80!black}{$\bm{9.45}_{\bm{\pm 0.15}}$} & \underline{$3006.11_{\pm 98.19}$} & \textcolor{red!80!black}{$\bm{9.63}_{\bm{\pm 0.17}}$} & \underline{$3151.85_{\pm 115.33}$} & \underline{$9.99_{\pm 0.17}$} & \underline{$3138.91_{\pm 109.00}$} & \underline{$9.98_{\pm 0.17}$} \\
FAIR & $2928.82_{\pm 102.35}$ & $10.31_{\pm 0.17}$ & $2964.76_{\pm 92.24}$ & $10.28_{\pm 0.16}$ & $2924.84_{\pm 111.93}$ & $10.27_{\pm 0.16}$ & $2930.30_{\pm 115.74}$ & $10.27_{\pm 0.19}$ \\
\rowcolor{green!8} \textbf{POLO} & \textcolor{red!80!black}{$\bm{3359.06}_{\bm{\pm 110.00}}$} & \underline{$9.61_{\pm 0.15}$} & \textcolor{red!80!black}{$\bm{3290.76}_{\bm{\pm 100.20}}$} & \underline{$9.72_{\pm 0.17}$} & \textcolor{red!80!black}{$\bm{3419.36}_{\bm{\pm 108.67}}$} & \textcolor{red!80!black}{$\bm{9.51}_{\bm{\pm 0.17}}$} & \textcolor{red!80!black}{$\bm{3360.17}_{\bm{\pm 102.54}}$} & \textcolor{red!80!black}{$\bm{9.58}_{\bm{\pm 0.18}}$} \\
\bottomrule
\end{tabular}
\end{table*}
}

{
\setlength{\aboverulesep}{0pt}
\setlength{\belowrulesep}{0pt}
  \begin{table}[t]
  \centering
  \caption{Inference time (seconds).}
  \label{tab:large-deep-inference-time}
  \small
  \setlength{\tabcolsep}{5pt}
  \renewcommand{\arraystretch}{1.05}
  \begin{tabular}{lccc}
  \toprule
  \rowcolor{HeaderGray}
  Method & $N_p=1$ & $N_p=2$ & $N_p=3$ \\
  \midrule
  TCAC & 4.0113 & 7.6692 & 9.1673 \\
  FAIR & 7.3350 & 9.0552 & 8.5202 \\
  \rowcolor{green!8} 
  POLO & \textcolor{red!80!black}{\textbf{3.9215}} & \textcolor{red!80!black}{\textbf{5.1377}} & \textcolor{red!80!black}{\textbf{3.6518}} \\
  \bottomrule
  \end{tabular}
  \end{table}
  }

\section{Related Work}
\noindent\textbf{Order Dispatch in On-Demand Services}.
Order dispatch is a fundamental problem in on-demand service systems and has been extensively studied over the past decades. Early approaches primarily relied on rule-based or heuristic strategies, such as greedy matching, collaborative filtering, or queue-based assignment~\cite{seow2009collaborative, zheng2018order, lee2004taxi}. While these methods are easy to deploy, they are often sensitive to parameter tuning and struggle to adapt to highly dynamic environments, where orders arrive continuously and supply-demand conditions evolve rapidly over time ~\cite{berbeglia2010dynamic}.
To improve adaptability, 
%
recent studies have increasingly explored learning-based methods for dispatch optimization. One line of work applies RL to optimize decision intervals or caching strategies, effectively decomposing the dynamic dispatch problem into a sequence of static optimization problems~\cite{li2022efficient, wang2019adaptive,li2019efficient}. Another line of research directly focuses on learning dispatch policies through RL, enabling end-to-end dispatch decision-making based on system states ~\cite{wang2025coopride, ma2021hierarchical,luo2023fleet}. Several studies combine these two directions; for example, Ma et al.~\cite{ma2021hierarchical} propose a hierarchical RL framework to jointly optimize dispatching and routing decisions in real-world supply chain systems. 
With the rise of instant delivery services, the concurrent order dispatch problem has been introduced to explicitly model overlapping delivery tasks~\cite{guo2021concurrent}. To address the increased complexity, subsequent work reformulated the RL modeling paradigm by shifting the agent perspective from service suppliers to service demanders and parameterizing heterogeneous couriers within the action space~\cite{zong2025deep}. Under this framework, Wang et al.~\cite{wang2023time} propose an actor-critic approach for food delivery dispatch under strict time constraints, while Jiang et al.~\cite{jiang2023faircod} further propose a competitive multi-agent framework to address income disparity among couriers. Despite their effectiveness in single-platform settings, most existing dispatch methods rely on strong assumptions, including full observability of couriers' states and exclusive platform control over dispatch decisions. These assumptions limit their applicability in realistic multi-platform environments, where couriers are shared across platforms and system states are only partially observable. 


\noindent\textbf{Crowdsourced Couriers and Multi-Platform Settings}.
Crowdsourced couriers and the coexistence of multiple platforms have become increasingly common in modern on-demand service markets ~\cite{savelsbergh2022challenges, alnaggar2021crowdsourced}, motivating growing research interest in the challenges arising from such settings.
A substantial body of work focuses on modeling crowdsourced worker behavior, including working-time patterns~\cite{mohri2025crowd},  task acceptance decisions ~\cite{ausseil2024online},and incentive modeling ~\cite{zheng2018order}. These studies provide important insights into individual courier behavior under dynamic and uncertain conditions. 
However, relatively little work explicitly studies order dispatch in multi-platform settings. A small number of studies~\cite{yang2022privacy,cheng2024cross,cheng2020real} investigate dispatch under multiple platforms with partial observation, typically distinguish between \emph{local} workers and \emph{cooperative} workers, and focus on privacy-preserving information sharing mechanisms. While these approaches relax some observability assumptions, they generally continue to assume exclusive courier affiliation or centralized control over dispatch decisions. In contrast, real-world instant delivery systems are characterized by decentralized decision-making and shared couriers, where each platform operates with platform-local information and without access to competitors' states. POLO directly addresses this setting by formulating multi-platform dispatch as a partially observable multi-agent learning problem, without requiring cross-platform coordination or information sharing. 


\noindent\textbf{Partially Observable MARL and Scalable Architectures}.
Multi-agent reinforcement learning (MARL) has been widely adopted in delivery and logistics optimization due to its ability to model the sequential, distributed decision-making inherent in large-scale dispatch systems~\cite{zong2022mapdp}.
A fundamental characteristic of MARL is that each agent operates under partial observability, receiving only a local observation rather than the full global state.
Existing work predominantly addresses spatial partial observability, where each agent's observation is limited to its local geographic region or neighborhood~\cite{zong2022mapdp}.
Such local observations introduce non-stationarity during training, as each agent's environment appears non-stationary from its local perspective when other agents simultaneously update their policies~\cite{hernandez2019survey}.
To address this, the Centralized Training with Decentralized Execution (CTDE) paradigm~\cite{lowe2017multi, rashid2020monotonic,yu2022surprising,foerster2018counterfactual} has emerged as a principled framework for partially observable MARL: a centralized critic with access to global state is used during training, while each agent executes its policy using only local observations.
%
%
However, these methods primarily treat partial observability as a spatial phenomenon and assume that a shared global state is accessible during centralized training.
In our work, we identify and formalize a more restrictive form of partial observability that arises in multi-platform instant delivery: beyond spatial locality, each platform agent is subject to a platform-specific information mask, whereby it can only observe couriers and orders belonging to its own platform.
This privacy-induced constraint applies equally during training, as aggregating per-platform states would require inter-platform coordination that is unavailable in practice, making the standard CTDE assumption inapplicable in this setting.

\section{Conclusion}

In this paper, we studied the order dispatch problem in realistic multi-platform instant delivery systems, where couriers are shared across platforms and each platform operates under partial observability. We propose POLO, a partially observable multi-agent reinforcement learning framework that enables decentralized dispatch using only platform-local observations. POLO combines an attention-based policy network to handle heterogeneous courier information with a counterfactual reward shaping mechanism to mitigate non-stationarity induced by joint actions across agents. Through extensive experiments, we demonstrated that POLO consistently improves platform revenue and reduces courier travel distance across varying system scales and numbers of platforms. This work provides a principled learning framework for multi-platform dispatch under partial observability and offers insights into the design of scalable and realistic decision-making systems for urban instant delivery. 

%

\section*{GenAI Disclosure Statement}
Generative AI tools were used solely for language refinement, grammatical correction, and figure-generation code assistance in this work.
\balance
\bibliographystyle{ACM-Reference-Format}
\bibliography{ref}
\end{document}